\documentclass[review]{elsarticle}

\usepackage{lineno,xcolor}
\usepackage{amsmath,amssymb}
\usepackage{siunitx} %The international system of units (SI)

\usepackage{enumitem}

\usepackage{wrapfig}
\usepackage{lscape}
\usepackage{rotating}
\usepackage{epstopdf}

\usepackage{tablefootnote}
\usepackage{soul}
\usepackage{subcaption} % this is for caption in subfigures
\usepackage{comment}

\usepackage[pagebackref=true,breaklinks=true,letterpaper=true,colorlinks,bookmarks=false]{hyperref}

\newcommand{\xdownarrow}[1]{%
  {\left\downarrow\vbox to #1{}\right.\kern-\nulldelimiterspace}
}

\begin{document}

\begin{frontmatter}

\title{Real-time Semantic Segmentation with  Context Aggregation Network}


\author{Michael Ying Yang, Saumya Kumaar, Ye Lyu, Francesco Nex}

% \cortext[cor1]{Joint first author}
% \cortext[cor1]{Joint corresponding author}

\address{Faculty of Geo-Information Science and Earth Observation (ITC), University of Twente, The Netherlands}

\begin{abstract}
With the increasing demand of autonomous systems, pixelwise semantic segmentation for visual scene understanding needs to be not only accurate but also efficient for potential real-time applications. 
In this paper, we propose Context Aggregation Network, a dual branch convolutional neural network, with significantly lower computational costs as compared to the state-of-the-art, while maintaining a competitive prediction accuracy. 
Building upon the existing dual branch architectures for high-speed semantic segmentation, we design a cheap high resolution branch for effective spatial detailing and a context branch with light-weight versions of global aggregation and local distribution blocks, potent to capture both long-range and local contextual dependencies required for accurate semantic segmentation, with low computational overheads. 
% Specifically, we achieve 76.6\% and 75.9\% mIOU on Cityscapes validation and test sets respectively, at 76 FPS on an NVIDIA RTX 2080Ti and 8 FPS on a Jetson Xavier NX. 
%Codes and training models will be made publicly available.
We evaluate our method on two semantic segmentation datasets, namely Cityscapes dataset and UAVid dataset. For Cityscapes test set, our model achieves state-of-the-art results with mIOU of 75.9\%, at 76 FPS on an NVIDIA RTX 2080Ti and 8 FPS on a Jetson Xavier NX. 
With regards to UAVid dataset, our proposed network achieves mIOU score of 63.5\%  with high execution speed (15 FPS).
\end{abstract}

\begin{keyword}
\texttt{Semantic segmentation \sep Real-time \sep convolutional neural network \sep context aggregation network}
\end{keyword}

\end{frontmatter}

\section{Introduction}
\label{sec:intro}

The last decade has witnessed the rapid development of scene understanding in the field of computer vision and photogrammetry, especially the fundamental semantic segmentation task.
Semantic segmentation is to assign labels for each pixel in the image, which has extensive applications, including scene understanding \cite{yang*17:scenegraph, YurenALYR20} , autonomous vehicles and driver assistance \cite{Geiger2012CVPR, Cordts2016Cityscapes}, augmented reality for wearables.
With the advance of deep learning \cite{krizhevsky_cnn_2012}, convolutional neural network (CNNs) are becoming the preferred approaches for semantic segmentation in recent years, e.g. FCN \cite{long_shelhamer_fcn_2015}, PSPNet \cite{zhao2017pyramid}.
These approaches produce highly accurate segmentation results, but often at the cost of reduced computational efficiency. 
For many applications, real-time performance is in fact often necessary, since semantic labeling is usually employed only as a preprocessing step of other time-critical tasks \cite{poudel2019fast}. 

%The real-time performance requires crucial trade-offs, especially in computationally intense designs, such as for pixel-wise image semantic segmentation. 

Low-latency semantic segmentation becomes a challenging task as the optimal balance between accuracy and efficiency, i.e. computational complexity, memory footprint and execution speed, is hard to achieve. 
Conventional real-time semantic segmentation architectures usually address only one of the above perspectives, thereby making high-accuracy designs computationally expensive and high-speed models relatively inaccurate. 
%Consider Table~\ref{table:citys} for instance, which displays a comparative performance of some of the fastest and the most accurate real-time segmentation architectures, benchmarked on the Cityscapes dataset \cite{Cordts2016Cityscapes}. From the table, we can observe that 
These high-speed models tend to have a relatively lower prediction accuracy, e.g.~\cite{poudel2018contextnet, poudel2019fast}, whereas the more accurate models tend to have lower inference speeds and higher computational overheads, e.g.~\cite{orsic2019defense, yu2018bisenet}.
There is a significant gap between the high-speed and high-accuracy architectures, in terms of computational expenses and execution speeds (See Table~\ref{table:citys}).
% \begin{figure}[h]
% 	\centering
% 	\includegraphics[width=\linewidth]{gg-1xx-min.jpg}
% 	\caption{Prediction results of our architecture on Cityscapes \cite{Cordts2016Cityscapes}. Leftmost column consists of the input image, middle column shows the prediction, whereas the last column shows the corresponding ground truth.}
% 	\label{fig:cab}
% \end{figure}

There are several challenges that are very commonly associated with real-time segmentation designs. Firstly, high-accuracy designs like \cite{orsic2019defense, yu2018bisenet} rely heavily on dense feature extractors such as ResNet-18 \cite{he2015deep}. Secondly, the shallow extractors utilized in the relatively high-speed algorithms such as \cite{poudel2018contextnet, poudel2019fast}, provide for lower computational expenses but are unable to extract sufficient features for accurate segmentation. Thirdly, even though the computationally expensive models are accurate, they suffer from some local and global inconsistencies during inference. These aforementioned inconsistencies are usually not found in non-real time methods like \cite{zhu2019asymmetric, li2019global}, but while designing low-latency architectures, the trade-offs are sometimes unfavourable. 
In these regards, inspired by popular dual-branch architectures, we propose a novel framework called Context Aggregation Network, where we design two branches, one for fast and effective spatial detailing and the other for dense context embedding. We further address the issue of local and global inconsistencies by reformulating global aggregation and local distribution (GALD) blocks \cite{li2019global} for real-time applications. 
Our speed-accuracy trade-offs and effective spatial-contextual feature fusion allow us to outperform the previous state-of-the-art approaches for real-time semantic segmentation 
on two public datasets, namely Cityscapes dataset \cite{Cordts2016Cityscapes} and UAVid dataset \cite{lyu2020uavid}. For Cityscapes dataset, we also evaluate our approach on an embedded device.

The remainder of the paper is structured as follows. 
In Sec.~\ref{sec:rel}, we review recent deep learning methods in both accurate and real-time semantic segmentation. 
Sec.~\ref{sec:method} introduces the architecture of our model with two branches.
In Sec. \ref{sec:exp}, we show our results on the Cityscapes dataset \cite{Cordts2016Cityscapes} and compare our model against other state-of-the-art models. 
Extensive ablation experiments are carried out on the Cityscapes dataset \cite{Cordts2016Cityscapes} to
evaluate our proposed method. 
We also test our model with the UAVid dataset \cite{lyu2020uavid}. 
Sec.~\ref{sec:con} concludes this paper.

\section{Related Work}
\label{sec:rel}

Semantic segmentation is one of the fundamental problems of computer vision. 
%The task aims at segmenting images and detecting various object categories within them. 
%Concretely, a scene parser classifies each pixel of an image into one of several predefined object classes.
%
A common approach before 2012 is to use pixel-based classifiers, such as random forests~\citep{Schroff*08:ObjectClassSegmentation},
and pixel-based conditional random fields (CRFs) \cite{lafferty*01:conditional} to improve performance by modelling neighbourhood dependencies. 
%A larger context is considered by so-called texton features. To handle feature selection among many potential features, \cite{TextonBoost2009} proposed a boosting scheme.
%
%Many approaches on semantic scene parsing address the combination of bottom-up cue and top-down cue
%\citep{toyoda*08:integration, plath*09:global, schnitzspan*09:discriminative}. They show promising results and specifically 
%improved performance compared to making use of only one type of cues. 
%Hierarchical random field models, which allow integration of features computed at different levels of 
%the hierarchy, have been proposed in \cite{plath*09:global,LadickyRKT14}.
%MAP inference in the model can be performed efficiently using graph cut based move making algorithms \cite{LadickyRKT14}.
%Joint object detection and image segmentation have been proposed in \cite{wojek*08:dynamic,gould*09:object,bao*11:object}.
%\cite{alahari*10:Learning} proposed an efficient large margin piecewise learning method for a CRF model, which reduces to an equivalent convex problem with a small number of constraints.
%The aforementioned learning procedure for these models is supervised learning.
%We have experience with supervised learning schemes and different inference methods as basis for interpretation with CRFs.
%
On the other side, {CNNs} are driving advances in computer vision in recent years, such as image classification \citep{krizhevsky_cnn_2012}, detection \citep{zhang_rcnn_2014}, recognition \citep{agrawal_nn_2014,oquab_cnn_2014}, semantic segmentation \citep{long_shelhamer_fcn_2015}, pose estimation \citep{toshev_pose_2014}.
%CNNs have shown remarkable performance in the large-scale visual recognition challenge (ILSVRC2012). 
%The success of CNNs is attributed to their ability to learn rich feature representations as opposed to hand-designed features used in previous methods. 
%
A number of recent approaches, including Recurrent CNN \cite{pinheiro2014}  and FCN \citep{long_shelhamer_fcn_2015}, have shown a significant boost in accuracy by adapting state-of-the-art CNN based image classifiers to the semantic segmentation problem.
\cite{pinheiro2014} presented a feed-forward approach for scene labelling based on Recurrent CNN. The system is trained in an end-to-end manner over raw pixels, and models complex spatial dependencies with low inference cost.
\cite{long_shelhamer_fcn_2015} has addressed the coarse-graining effect of the CNN by upsampling the feature maps in deconvolution layers, and combining fine-grained and coarse-grained features during prediction. 
% This approach, combining down-sampling with subsequent up-sampling, is necessary to maintain a large receptive field without increasing the size of the convolution kernels, which otherwise become difficult to learn. 
%
% Recent work in this area has attempted to combine CNNs with CRFs to achieve state-of-the-art results on many datasets.
%\cite{FarabetCNL13} use a multiscale CNN trained from raw pixels to extract dense feature vectors that encode regions of multiple sizes centered on each pixel.
%Output of this CNN serves as unary potential in a pairwise CRF for scene labeling.
\cite{ChenPKMY15} proposed a framework to minimize coarse-graining by skipping multiple sub-sampling layers and avoid introducing additional parameters by using sparse convolutional kernels in the layers with large receptive fields, with a dense CRF \citep{KrahenbuhlK11} as post-processing step. 
\cite{crfasrnn_iccv2015} has addressed coarse-graining by expressing a mean-field CRF with Gaussian pairwise potentials as a Recurrent CNN, and concatenating this Recurrent CNN behind a FCN, for end-to-end training of all parameters.

Real-time semantic segmentation has been addressed using diverse approaches. Romera \textit{et al.} \cite{romera2017erfnet} proposed to use factorized convolutions with residual connections for maintaining a balance between accuracy and execution speed. Poudel \textit{et al.} \cite{poudel2018contextnet} suggest a dual-branch network with bottlenecks to effectively capture local and global context for fast segmentation. Later they propose an improved learning-to-downsample module in \cite{poudel2019fast} for improved trade-offs between execution speed and accuracy.   Accurate dual-branch segmentation networks were suggested by Yu \textit{et al.} \cite{yu2018bisenet}, where novel feature fusion and attention refinement modules for accurate semantic segmentation tasks were proposed. Multiple encoder-decoder pairs with multi-scale skip connections were also studied in this regard in \cite{zhuang2019shelfnet}. This ensemble of shallow and deep paths is viewed as a shelf of multiple networks allows for effective feature representation with shallower backbones like ResNet-34, as compared to \cite{yu2018bisenet, zhao2017pyramid}. Another approach to real-time segmentation is by using depth-wise asymmetric bottlenecks \cite{li2019dabnet}, which theoretically provides for a sufficient receptive field as well as captures dense context.

Attention modules have the capability to model long-range dependencies, and several authors have employed the concept of attention in various works \cite{lin2016efficient, lin2017structured, vaswani2017attention, shen2018disan}. The introduction of attention to machine understanding was achieved first in \cite{lin2017structured}, where the global dependencies of inputs were learnt, which were then applied to natural language processing. Since then, a lot of works have utilized this concept for several scene understanding tasks at both single and multiple scales \cite{fu2019dual, li2019global, ramachandran2019stand, zhong2020squeeze, zhu2019cross, tao2020hierarchical, huang2019interlaced}, thereby outperforming the previous conventional context embedding methodologies.

Another context-focused work was published by Jiang \textit{et al.} \cite{jiang2019context} where they introduced context refinement and context integration modules for efficient scene segmentation. Light-weight feature pyramid encoding models were suggested in \cite{liu2019feature}, which is an adaptation of the regular encoder-decoder architecture with depth-wise dilated convolutions. Multi-scale context aggregation was presented in yet another couple of approaches \cite{si2019real, zhang2020farsee}, where \cite{si2019real} uses class boundary supervision to process certain relevant boundary information and \cite{zhang2020farsee} use optimized cascaded factorized ASPP \cite{chen2017deeplab} module to balance the trade-offs between accuracy and execution speed. 
Orsic \textit{et al.} \cite{orsic2019defense} developed an approach which exploits light-weight upsampling and lateral connections with a residual network as the main recognition engine for real-time scene understanding. This particular algorithm is deemed as the current state-of-the-art network for real-time semantic segmentation on Cityscapes dataset.

\section{Method}
\label{sec:method}

\begin{figure}[http]
	\centering
	\includegraphics[width=1\linewidth]{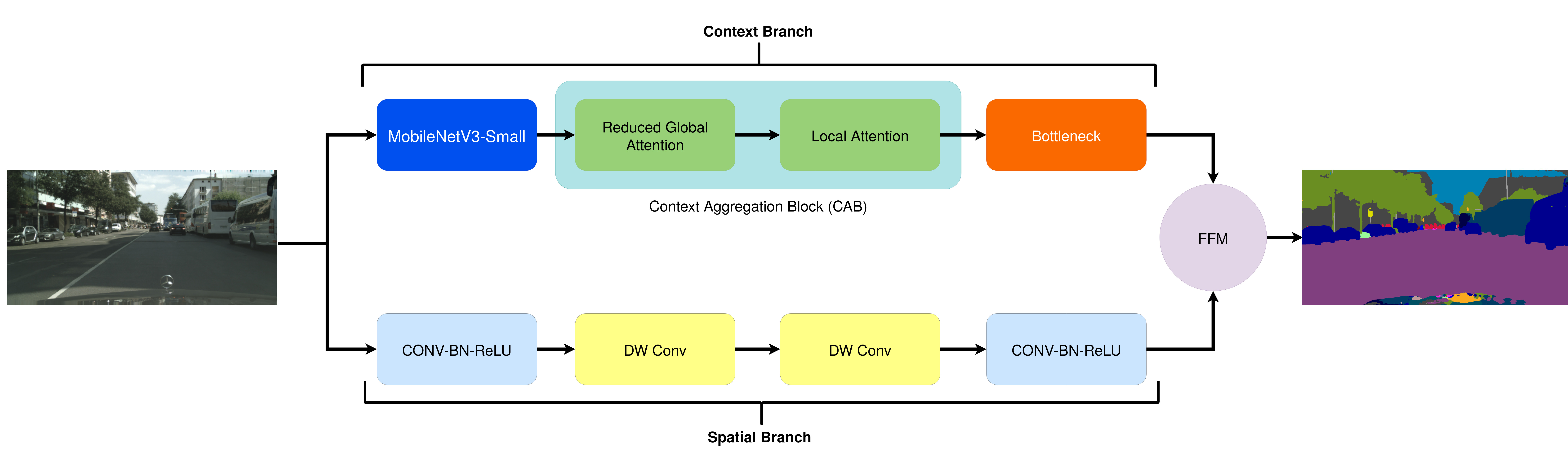}
	\caption{Overall model architecture of Context Aggregation Network. The spatial and context branches allow for multi-scale feature extraction with significantly low computations. Fusion block (FFM) assists in feature normalization and selection for optimal scene segmentation. The bottleneck in the context branch allows for a deep supervision into the representational learning of the attention blocks.}
	\label{fig:cabinet}
\end{figure}

We illustrate the architecture of our Context Aggregation Network in Fig.~\ref{fig:cabinet} with two branches, one for fast and effective spatial detailing and the other for dense context embedding. The spatial and context branches allow for multi-scale feature extraction with significantly low computations. 
These two branches are then fused in the fusion block (FFM) for the final object category prediction. 
Each component of our model is described in details in the following sections. 

\subsection{Spatial Branch}
\label{sec:spatial}

In order to encode sufficient spatial information, multiple existing approaches \cite{peng2017large,wang2018understanding,chen2017rethinking,chen2017deeplab} have employed the usage of dilated convolutions, while others attempt to capture large receptive fields with either pyramid pooling or large-sized kernels \cite{peng2017large,chen2017rethinking,zhao2017pyramid}. These approaches indicate that sufficient receptive fields and effective spatial information encoding, could be crucial for accurate semantic segmentation. 
It is however, difficult to satisfy both the requirements in parallel, especially when designing real-time segmentation architectures. 

Conventional real-time designs usually either downsize the image to a smaller resolution \cite{zhao2018icnet} or use a lightweight reduction model \cite{badrinarayanan2017segnet,paszke2016enet} for speeding up the overall architecture. Downsizing the image however, incurs a loss in the spatial information and light-weight reduction models like Xception \cite{chollet2017xception} tend to damage the receptive fields because of the incessant channel pruning, especially in the early stages \cite{badrinarayanan2015segnet, paszke2016enet}. This problem was addressed in \cite{yu2018bisenet}, but at the cost of significant increase in the computations, thereby imparting a lower execution speed on mobile and embedded platforms. Based upon these observations, we propose a shallow branch that encodes rich spatial information and maintains an adequate receptive field, while maintaining a significantly low computational cost from a full-resolution image. 
Specifically, this path has four convolutional layers, where the first layer has a large kernel size, followed by batch normalization and ReLU, followed by two depth-wise convolutional layers. A strategic use of depth-wise convolutions results in the same outcomes as that of conventional convolutions, but with reduced computations, and the marginal loss in features can be compensated by enlarging the number of feature representations. Finally, the last layer is another convolutional layer with kernel size of 1. Strides for the first three layers are fixed at 2, whereas the last layer has a unit stride. This branch (Spatial Branch in Fig.~\ref{fig:cabinet}), hence generates an output that is $\frac{1}{8}$ of the input resolution \cite{yu2018bisenet}, thereby maintaining the required spatial information with a significant reduction in computations.

\subsection{Context Branch}

Detailed spatial information coupled with adequate receptive field significantly affects semantic segmentation accuracy \cite{yu2018bisenet}. While the spatial branch in Sec.~\ref{sec:spatial} takes care of the spatial details, we design a new attention-based context branch, with light-weight global aggregation \cite{li2019global} and local attention \cite{li2019global} blocks, for providing a sufficient receptive field and capturing both global and local context. We use a pre-trained MobileNetV3-Small \cite{howard2019searching} as the lightweight feature extractor in this branch, which can downsample the input images effectively and efficiently, to provide rich high level semantic features. These features are however unrefined, and hence, need to be passed on to a refinement stage, termed as the context aggregation block comprising of reduced global attention \cite{li2019global} and local attention sub-modules.

\subsubsection{MobileNetV3-Small}

MobileNetV3-Small \cite{howard2019searching} employs a mixture of layers suggested in MobileNetV2 \cite{sandler2018mobilenetv2} and MnasNet \cite{tan2019mnasnet}, to construct the most effective and efficient neural network for mobile applications. Modified \verb|swish| non-linear functions were used to improve the performance of layers, along-with hard sigmoid for squeeze-and-excitation modules. 
% The final feature vector after the backbone is of size $64\times 32\times 576$, for a given Cityscapes \cite{Cordts2016Cityscapes} input shape of $2048\times \times 3$. The network is however, fully convolutional and supports arbitrary input sizes.
The network is fully convolutional and supports arbitrary input sizes.
We use a pre-trained MobileNetV3-Small \cite{howard2019searching} as the lightweight feature extractor, which can downsample the input images and provide the semantic features.

\subsubsection{Context Aggregation Block}

The long-range and local dependencies in the representational outputs of feature extractors are crucial for accurate semantic segmentation \cite{zhu2019asymmetric, li2019global, wang2018non}. The proposed context aggregation block is to capture such inter-channel and intra-channel mappings, effectively and efficiently. 
Several previous works have suggested modules to effectively acquire such semantics \cite{li2019global, fu2019dual, zhu2019asymmetric, li2019attention}. For our work, 
We adopt the global attention (GA) block from global aggregation and local distribution (GALD) \cite{li2019global}. This module is potent enough to capture long-range dependencies crucial for accurate semantic segmentation but is computationally expensive and requires significant GPU memory for execution. 
%In the next sub-section, we review the shortcomings the components of GA block and attempt to overcome them.

\begin{figure}[t!]
	\centering
	\includegraphics[width=\linewidth]{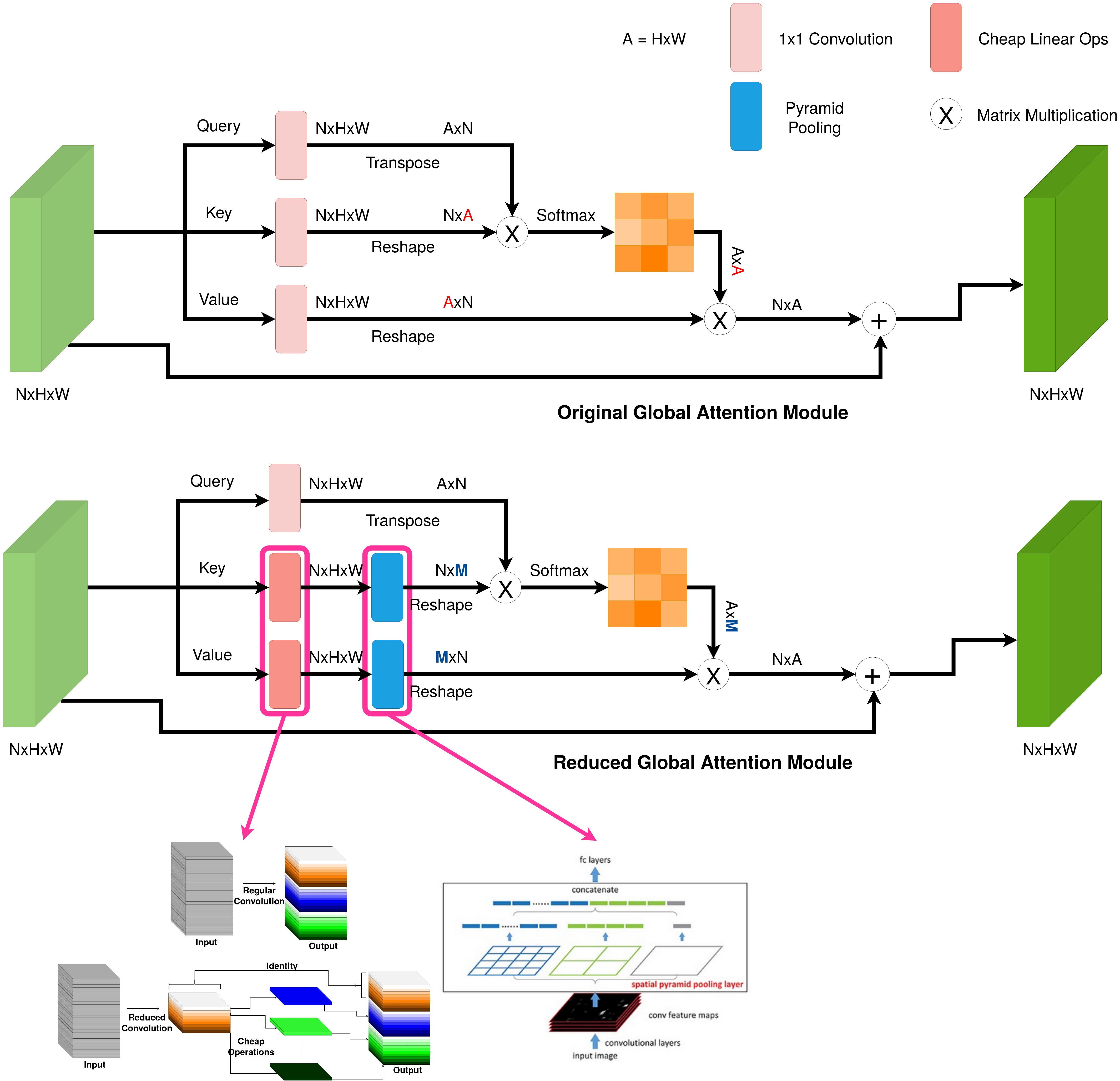}
	\caption{Original global attention module (Top) and reduced global attention module (Bottom). 
	Our reduced GA module leverages the benefits of spatial pyramid pooling (SPP) \cite{zhao2017pyramid} and depth-wise separable convolutions.
	SPP module allows for a strategic selection of representative features, whereas the cheap linear operations (CLO) concept leads to use multiple kernel sizes within a single convolution.}
	\label{fig:cab}
\end{figure}

\noindent
\textbf{Reduced Global Attention Block.}
Fig.~\ref{fig:cab} Top shows the flow of the original global attention (GA) module \cite{li2019global}, where the output from the previous backbone stage is fed to three parallel convolutional layers to generate new embeddings. 
After the embeddings have been generated similarity matrices are calculated using matrix multiplications, followed by a Softmax normalization process. 
The output contains the semantic cues for every position in the input feature vector. 
There are two limitations to this GA module \cite{li2019global} for real-time semantic segmentation. 
Firstly, the original design proposes to extract the contextual information directly from the outputs of the backbone using three parallel convolution layers, thereby increasing the required number of parameters. Secondly, the matrix multiplications of the \verb|Key| and \verb|Value| convolutions followed the by the next multiplication process after the softmax activation stage, increase the time complexity, as these computations are performed on relatively large matrices \cite{zhu2019asymmetric}.

The matrix multiplications are large because of the size of the input feature vector, $A$, and if it were changed to a smaller value $M$, (where $M << A$), it might help in alleviating some of the computations. Although, changes have to be made in such a way that the output size of the vector remains unchanged. Hence, we employ spatial pyramid pooling (SPP) modules \cite{zhao2017pyramid} in the global attention module to effectively reduce the size of the feature vectors (Fig. \ref{fig:cab} Bottom). Instead of feeding all the spatial points to the multiplication process, it becomes more feasible to sample the points and feed only certain representative points to the process. Following \cite{zhu2019asymmetric}, we use four adaptive maximum pooling at four scales and the pooling results are flattened and concatenated to serve as the input to the next layer. 
For our experiments, the number of sparse representations can be formulated as $M = \sum_{n \in {1, 3, 5, 8}} n^2 = 110$, thereby reducing the complexity to $\mathcal{O}(\hat{N}AM)$, which is much lower than $\mathcal{O}(\hat{N}A^2)$. 
For example, for the input to the GA block of $64 \times 32 = 2048$, this asymmetric multiplication saves $\frac{64 \times 32}{110} \approx 18$ times the computation cost. Furthermore, the feature statistics captured by the pooling module are sufficient to provide cues about the global scene semantics.

Next, this block employs three parallel $1\times 1$ convolution layers, which results in a relatively larger number of parameters. This might not have a direct influence on the overall execution speed, but a neural design with lesser parameters demonstrates the efficiency of the model. Regular convolution layers have multiple learnable filters that convolve on the input feature vector. It was suggested in \cite{han2020ghostnet} that these regular convolution layers can be replaced with a concept called cheap linear operations (CLO), which is graphically depicted in Fig. \ref{fig:cab} Bottom. 
The key idea is to begin with a smaller convolution, and then perform a set of linear transformations on the output of the smaller convolution to generate new representation. Then to simply stack the representations from both the stages to create a set of features that correspond to a full convolution operation. These linear transformations significantly reduce the parameters and computations \cite{han2020ghostnet} as compared to regular convolutions. 
%A quantification of these improvements have been presented in Table \ref{table:pam-cap}.

\noindent
\textbf{Local Attention.}
The global statistics for every group in the GA module are later multiplied back to features within. However, the windows in which the statistics are calculated are relatively large, and hence there is a possibility that the statistical cues could be biased to towards the larger patterns as there are more samples within, which can further cause over-smoothing of the smaller patterns.

In this regard, a local attention (LA) module was proposed in \cite{li2019global} to adaptively use the features, considering patterns at every position encoded by the previous global attention block. LA module is adapted directly after the GA block as a fine-tuning stage for the global semantic cues. 
Our ablation studies indicate that this module is efficient and fast. 
The LA block predicts local weights by re-calculating the spatial extent, which is primarily targeted to avoid coarse feature representation issues present in the previous GA module. Here, the predicted local weights add a point-wise trade-off between the global information and local context. Therefore, the local attention block is modelled as a set of three depth-wise convolutional layers, which allows for fine-tuning the feature representations from the previous GA module.

\subsubsection{Bottleneck}

Inspired from previous works \cite{howard2019searching, he2015deep}, we design a simple downsampling module to restrict the representation of the refined features in the depth dimension. Furthermore, this restriction, when adapted directly post the context aggregation stage allows us to supervise the representational learning of the attention blocks and context branch.

\subsection{Feature Fusion Module (FFM)}

\begin{figure}[http]
	\centering
	\includegraphics[width=1\linewidth]{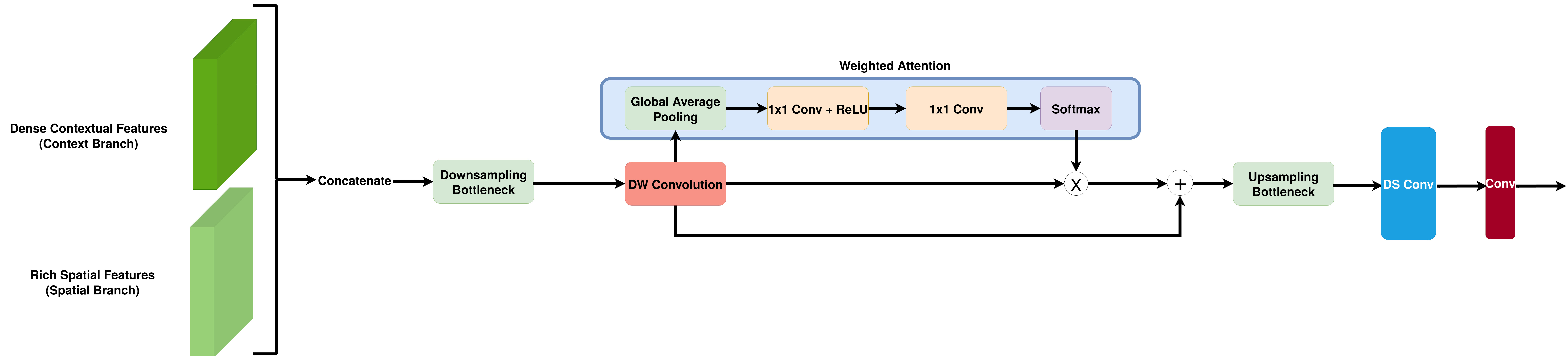}
	\caption{Feature fusion module.}
	\label{fig:fusion}
\end{figure}

It is to be noted that the features extracted from both the branches are at different scales of representation and require a scale normalization for effective fusion. Hence, a simple addition of both features \cite{poudel2018contextnet, poudel2019fast} to save computations, is unlikely to produce desirable accuracy. Therefore in this work, we implement a feature fusion technique as suggested in \cite{yu2018bisenet}, with certain adaptations. In order to fully utilize the vector representations from both the branches, we concatenate both the features first, followed by a downsampling bottleneck. After the concatenation, the final feature representation has large dimensions, which increases the amount of required computations. Adding a downsampling bottleneck reduces these computations in the later stages of feature selection (weighted attention) by a significant margin, without causing damage to the overall accuracy (see Table \ref{table:ffm_variants}). The weighted attention section inspired from \cite{yu2018bisenet, hu2018squeeze} is added to selectively weigh the features in terms of their contribution to the overall prediction accuracy. These selected features are later upsampled to generate the same number of representations as \cite{yu2018bisenet}, but with significant reduction in computations. 
The final two layers after the upsampling bottleneck generate the final output predictions. We use only two layers in this case because for a simple class-wise separation, multiple layers become unnecessary, hence one depth-wise separable convolution and one point-wise convolution are sufficient. A detailed schematic is shown in Fig. \ref{fig:fusion}.

\subsection{Loss Functions}

For training the Context Aggregation Network, we use three cross entropy loss functions with online hard example mining \cite{shrivastava2016training}, one (primary) for the final output and two (auxiliary) for the context branch. 
The auxiliary loss functions allow for a deep supervision of the learning of the context branch and attention modules. %Individual CE loss functions can be defined as:
% \begin{equation}
% CE_{loss}=\frac{1}{N} \sum_{i} L_{i}=\frac{1}{N} \sum_{i}-\log \left(\frac{e^{p_{i}}}{\sum_{j} e^{p_{j}}}\right)
% \end{equation}
% where, $p$ is the final output of the network (prediction). 
The overall joint loss representation of our model $L(X ; W)$ can be formulated as:
\begin{equation}
L(X ; W) = l_{p}(X ; W) + l_{c1}(X_1 ; W) + l_{c2}(X_2 ; W) 
\end{equation}
where, $l_p$ is the principal loss for monitoring the overall output, $l_{c1}$ is the auxiliary loss for the reduced global attention module of the network, $l_{c2}$ is the auxiliary loss for the local attention module, $W$ are the network parameters, and $p$ is the final output of the network prediction. 
Utilizing a joint loss makes it easier to optimize the model, as suggested in \cite{zhu2019asymmetric, yu2018bisenet}. 
%However, the auxiliary losses are only employed during training and not in the inference stage.
%Apart from just monitoring the overall output of CABiNet, we use two additional auxiliary loss functions, one that monitors the output of the attention branch and one for the attention fusion module. These two auxiliary loss functions provide deep supervision of the two modules.

\section{Experiments}
\label{sec:exp}

In this section, experiments are shown to evaluate the effectiveness of the proposed Context Aggregation Network on two publicly available datasets. 
We compare the performance of our model against that of other the
state-of-the-art models on the Cityscapes dataset \cite{Cordts2016Cityscapes}. 
We also conduct a comprehensive ablation study on the Cityscapes dataset \cite{Cordts2016Cityscapes} to show the effectiveness of our proposed method and evaluate how network structure influence the model performance. 
Next the UAVid dataset \cite{lyu2020uavid} is used to further demonstrate the advantages of our method.

\subsection{Experiments on Cityscapes dataset}

\subsubsection{Dataset}
We benchmark our proposed approach on the Cityscapes dataset~\footnote{https://www.cityscapes-dataset.com/} \cite{Cordts2016Cityscapes}. 
% Cityscapes is an urban-scene understanding dataset which contains a total of fully annotated 5000 images out of which, 2975 are for training, 500 for validation and the remaining 1525 for testing. This dataset contains 35 classes, out of which 19 are used for urban scene understanding and the image size is 1024$\times$2048. 
Cityscapes is an urban scene understanding dataset. The dataset contains a total of 5000 images (fine-grained) out of which, 2975 are for training, 500 for validation and the remaining 1525 for testing. 
The dataset also contains additional 20K coarsely annotated images, but we do not use them in this work. 
The image size for this dataset is 1024$\times$2048, collected from across 50 different cities. The densely annotated data contains 35 classes, out of which 19 are used for urban scene understanding. 
% Usage of coarse training data could be employed but usually it is not a practise for real-time low-latency architectures. 
% An example image and the corresponding ground truth are shown in Fig. \ref{fig:example-citys}.

\subsubsection{Training Setting}
For optimizing the network, we use Stochastic Gradient Descent (SGD) \cite{kiefer1952stochastic} and set the initial learning rate as $e^{-4}$ for Cityscapes. We employ the poly-learning rate strategy, where during training, the learning rate is multiplied with $1 - (\frac{iter}{max\_iter})^{power}$, with power being equal to 0.9. For Cityscapes image, we randomly crop patches of [1024, 1024] from the original input images during training. We use data augmentation techniques like random horizontal flips, random scaling and color jitter. 
Scales range from (0.75, 1.0, 1.5, 1.75, 2.0). The batch size is set at 6 for Cityscapes and training iteration is set at 160k. All the experiments are conducted on a single NVIDIA RTX 2080Ti and Jetson Xavier NX, with PyTorch.

\subsubsection{Evaluation metrics}

For evaluation, we use the standard mean of class-wise intersection over union (mIOU), memory footprint (in MB/GB), GLOP count (floating point operations) and the overall execution speed (Frames per second). For calculating the computational expenses we adopt the same strategies as mentioned in \cite{orsic2019defense, yu2018bisenet}.

\subsubsection{Comparison with state-of-the-art}

A detailed comparison between our Context Aggregation Network and other architectures has been provided in Table \ref{table:citys}, based upon the GPU memory footprint, MAdd/GLOPs count, execution speed (RTX 2080Ti) and the overall mIOU score on the Cityscapes validation and test sets. 
As it can be observed from the table, our model outperforms the previous methods for real-time scene understanding and achieves the highest mIOU scores of 76.6\% and 75.9\% on validation and test sets respectively. 
In comparison with the most memory efficient model SINet \cite{park2020sinet}, our model has 7.7\% higher mIOU and 7.8 FPS faster speed.  
In comparison with fastest model Fast-SCNN \cite{poudel2019fast}, our model has  7.5\% higher mIOU.  
In comparison with the most recent work GAS \cite{lin2020graph}, our model has  4.1\% higher mIOU, while being competitive regarding the speed assuming full resolution image input. % 
Note that the FPS number of GAS \cite{lin2020graph} was computed on a smaller image input of 769$\times$1537.

Furthermore, comparing our proposed method with the previously established state-of-the-art algorithms \cite{poudel2019fast, orsic2019defense, yu2018bisenet}, our improvements favour both accuracy and speed simultaneously. Computational overheads such as parameter count, GFLOPs etc. in our architecture are significantly lower than the existing accurate real-time architectures, with increased accuracy. Optimized GALD-blocks coupled with efficient spatial detail and light-weight dense extractors, allow our approach to outperform the conventional real-time semantic segmentation architectures in multiple aspects. 

\begin{figure}
\centering
	\includegraphics[width=0.98\textwidth]{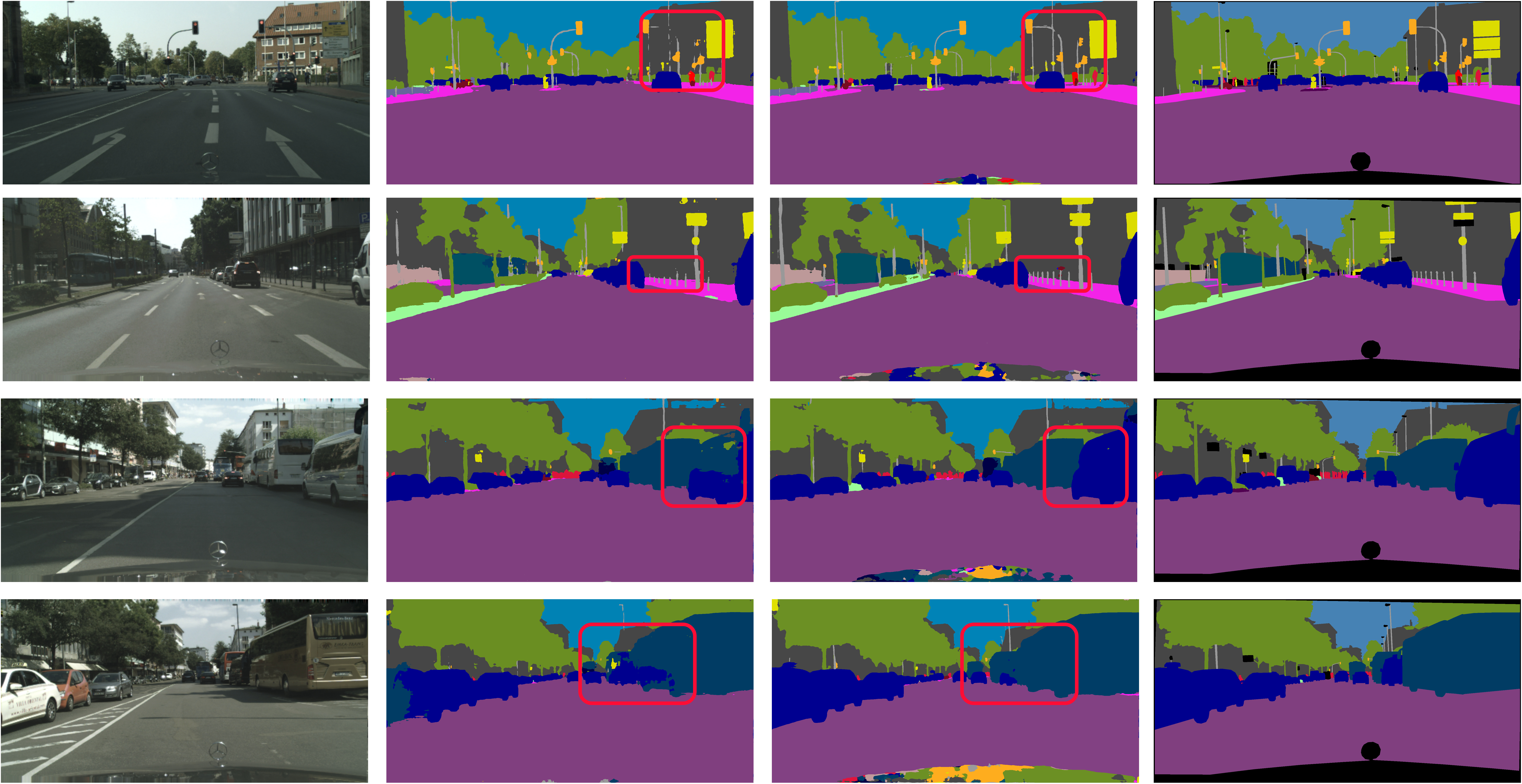}
	\caption{Semantic segmentation results on the Cityscapes validation set. From left, the first column consists of the input  images; the second column indicates the prediction results of SwiftNet~\cite{orsic2019defense}; the third column shows the predictions from our architecture and the red boxes show the regions of improvements and the last column comprises of the ground truths.}
	\label{fig:cabinet-citys}
\end{figure}

\begin{figure}
\centering
	\includegraphics[width=0.98\textwidth]{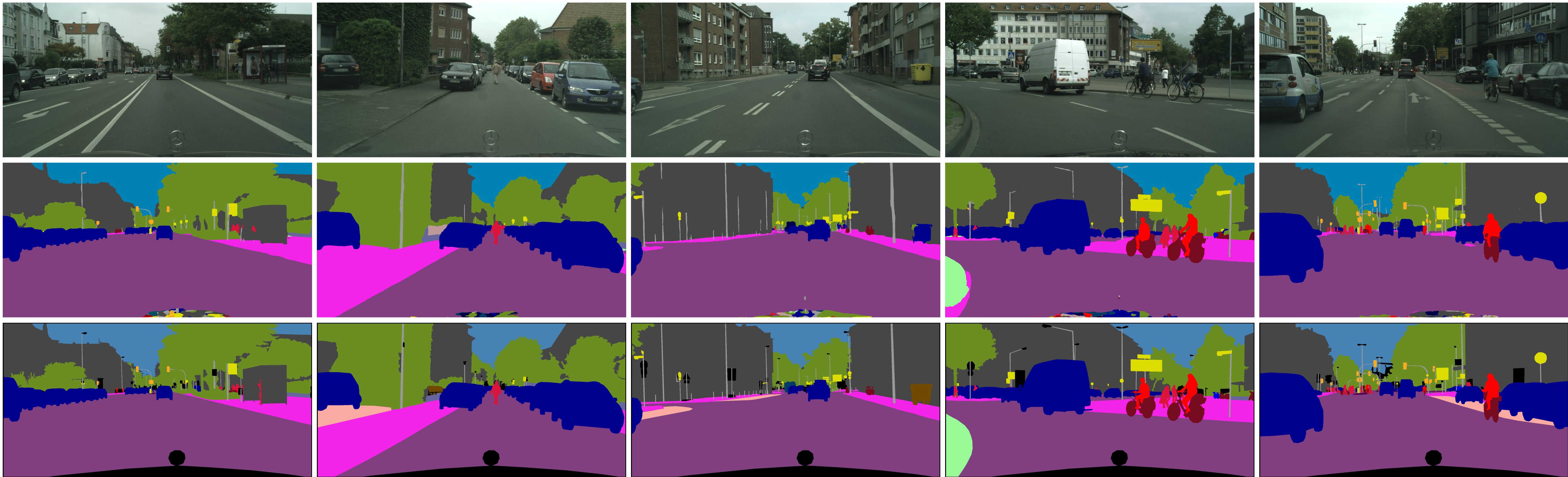}
	\caption{More semantic segmentation results on the Cityscapes validation set. 
	First row consists of the input RGB images. Second row contains
the predictions from our model and the third row shows the ground truths of the input images.}
	\label{fig:cabinet-citys-val}
\end{figure}

\begin{figure}
\centering
	\includegraphics[width=0.98\textwidth]{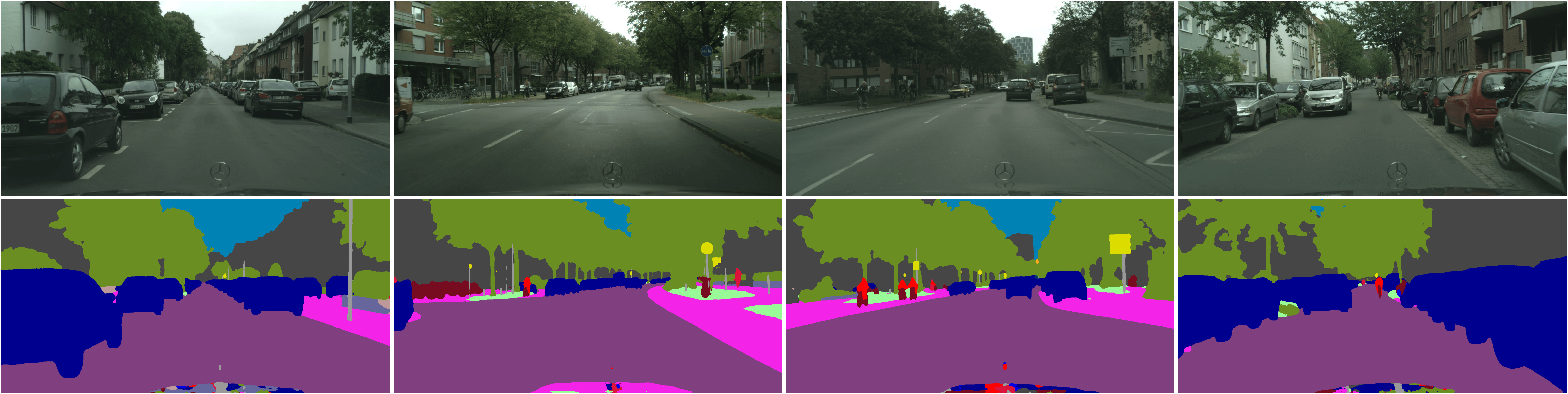}
	\caption{Semantic segmentation results on the Cityscapes test set. Top row shows the images, and the bottom row shows the predictions.}
	\label{fig:cabinet-citys-test}
\end{figure}

Qualitative results are shown in Fig.~\ref{fig:cabinet-citys}. Compared with \cite{orsic2019defense}, our model has better performance in terms of detecting under-represented objects like poles, traffic signs, etc. Thanks to the efficient global and local semantic aggregation, our model does not suffer from such local or global inconsistencies. 
More qualitative results on Cityscapes validation set are shown in Fig.~\ref{fig:cabinet-citys-val}. 
Qualitative results on Cityscapes test set are shown in Fig.~\ref{fig:cabinet-citys-test}.

\begin{table}[htp!]
\centering
\resizebox{1.1\textwidth}{!}{  
\begin{tabular}{lccccccc|c}
\hline
\textbf{Model}& \multicolumn{2}{c}{\textbf{mIOU}}  & \textbf{Memory} & \textbf{MAdd}& \textbf{Flops}  & \textbf{Params} & \textbf{FPS} & \textbf{FPS*}\\ 
\cline{2-3}
 & \textbf{val} & \textbf{test} & \\ \hline
CGNet \cite{wu2018cgnet}  & --&64.8 & 3134.91MB  & 55.01G & 27.05G & 0.5M & 34.91&  2.91\\
ContextNet \cite{poudel2018contextnet}& --&66.1 & 1429.43MB  & 13.98G & 6.74G& 0.88M& 118.65  &  10.49\\
%LiteSeg \cite{emara2019liteseg} & --& 67.8 & --  & --& -- & -- & --&  --\\
%FPENet \cite{liu2019feature} & 69.5&68.0 & --  & --& -- & -- & --&  --\\
SINet \cite{park2020sinet}& 69.4&68.2 & \textbf{672.00MB}& \textbf{2.99G}& \textbf{1.24G}& \textbf{0.12M}& 68.61&  12.02 \\ 
Fast-SCNN \cite{poudel2019fast}& --&68.4 & 1239.33MB  & 13.85G & 6.72G& 1.14M& \textbf{128.97}&  \textbf{11.49}\\ 
%FarSee-Net \cite{zhang2020farsee}  & 69.8 & 68.4 & --  & --& -- & -- & --&  --\\
%ERFNet \cite{romera2017erfnet} & 71.5 & 69.7 & 3244.00MB  & 213.88G& 102.61G& 2.07M& 20.46&  1 \\ 
%DABNet \cite{li2019dabnet} & 70.1 & 70.0 & 3287.50MB  & 82.83G & 40.88G & 0.76M& 40.35& 1 \\ 
DFANet \cite{li2019dfanet} & 71.3 & 70.1 & 1778.09MB  & 30.68G & 15.28G & 2.19M& 47.88&  4.71\\ 
LedNet \cite{wang2019lednet}  & 71.5 & 70.6 & 3031.75MB  & 90.71G & 45.84G & 0.93M & 24.72& 0.7\\
ESNet \cite{lyu2019esnet} & --&70.7 & 1176.29MB  & 66.81G & 33.81G & 1.81M & 55.65&  4.65\\
ShelfNet \cite{zhuang2019shelfnet} & 75.2 & 74.8 & 1158.12MB  & 187.37G& 93.69G & 14.6M& 44.37 &  2.59 \\
SwiftNet \cite{orsic2019defense} & 75.4 & 75.5 & 1671.66MB  & 207.64G& 103.37G& 11.80M & 45.40&  2.61 \\
BiSeNet \cite{yu2018bisenet}& 74.8 & 74.7 & 1941.39MB  & 208.18G& 103.72G& 12.89M & 65.50 &  2.42\\  
GAS \cite{lin2020graph}  & 72.4 & 71.8 & --  & --& -- & -- & 108.40&  --\\ \hline 
Ours& \textbf{76.6} & \textbf{75.9} & 1256.18MB  & 24.37G & 12.03G & 2.64M& 76.50 &  8.21 \\ \hline
\end{tabular}
}
\caption{Comparison with state-of-the-art on the Cityscapes dataset \cite{Cordts2016Cityscapes}. 
For all the network models, mIOU are taken directly from the original publications.
FPS indicates the average model run-times on a single RTX 2080Ti with an input resolution of 1024$\times$2048.  
We recompute the MAdd and FLOPS on the full resolution image from the official implementations, except GAS \cite{lin2020graph}.
Note that FPS* indicates the average model run-times on a single Jetson Xavier NX  with an input resolution of 1024$\times$2048 (see Sec.~\ref{sec:jetson}).  
}
%For most models, we recompute the MAdd and FLOPS on the full resolution image from the official implementations, except GAS \cite{lin2020graph}, where $-$ indicates that the corresponding values were not available. 
%Note that the reported FPS number of BiSeNet in \cite{yu2018bisenet} was 65.5, and the FPS number of GAS \cite{lin2020graph} was computed on a smaller image input of 769$\times$1537.}
\label{table:citys}
\end{table}

\begin{table}[h!]
\centering
\begin{tabular}{lcccccccc}
\hline
\textbf{Model}& \textbf{mIOU }& \textbf{FPS}  \\ \hline
Baseline & 68.4 & 110.65\\
Baseline + SB + CB  & 72.3 & 86.76\\ 
Baseline + SB + CB + CAB & 74.7 & 81.20\\
Baseline + SB + CB + CAB + FFM & 76.6  & 76.50\\ \hline
\end{tabular}
\caption{Basic ablation study. SB and CB stand for spatial and context branches, whereas FFM stands for feature fusion module respectively. } %FPS has been indicated on a single RTX 2080Ti.
\label{table:ablation}
\end{table}

\subsubsection{Ablation Studies}

Baseline is defined as a simple dual-branch network with two convolution layers in the spatial branch and untrained feature extractor in the second branch. The baseline is devoid of attention and bottleneck modules and is similar in structure with \cite{poudel2018contextnet}. For fusing the features from both branches, we simply add them which are later discriminated by a small classifier block into the respective number of classes. Both branches are fed images at the same resolution, unlike \cite{poudel2018contextnet} and all the ablation experiments are performed on this baseline.
In order to clarify how different modules contribute to the performance, we ablate different components and present the results in Table \ref{table:ablation}.
%Basic ablation study. SB and CB stand for spatial and context branches, whereas FFM stands for feature fusion module respectively

\paragraph{\textbf{Context Aggregation Block}}
The context aggregation block (CAB) is designed specifically to capture local and global context effectively and efficiently. If we remove CAB from the design keeping all other modules and training/inference parameters intact, we observe a drop of 2.1\% in the overall mIOU score, along with a drop in inference time by almost 3ms. This implies that the addition of the context block enhances the feature representations, while having minimal impact on the overall execution speed and complexity. Table \ref{table:others_with_cab} further proves the efficacy of the context aggregation block, which can be used as a plug\&play module with other dual-branch architectures for semantic segmentation.

\begin{table}[h!]
\centering
\begin{tabular}{lcccccccc}
\hline
\textbf{Model}& \textbf{mIOU w/o CAB}  & \textbf{mIOU w CAB} \\ \hline
ContextNet \cite{poudel2018contextnet}& 66.1 & 69.2 \\
Fast-SCNN \cite{poudel2019fast} & 68.4 &  71.2 \\ 
BiSeNet \cite{yu2018bisenet}& 74.7 &  75.3 \\\hline
%BiSeNetV2 \cite{yu2020bisenet}& 74.2 &  75.1 $\uparrow$\\ 
\end{tabular}
\caption{CAB implemented in other algorithms. Straightforward addition to \cite{ poudel2018contextnet, poudel2019fast} results in significant improvements over the baseline models. In \cite{yu2018bisenet}, the attention refinement modules were replaced with CAB.}
\label{table:others_with_cab}
\end{table}

Interestingly, using SPP modules \cite{zhao2017pyramid} for attention modules was suggested in \cite{zhu2019asymmetric}, but adding cheap linear operations (CLO) \cite{han2020ghostnet} not only reduces the required computations, but also provides a slightly better accuracy (Table \ref{table:pam-cap}). This could be attributed to the fact within these linear transformations, there can be multiple kernel sizes  \cite{han2020ghostnet}, thereby allowing for multi-scale feature aggregation.

\begin{table}[ht]
\centering
\small
\begin{tabular}{lcccccccc}
\hline
\textbf{Module}& \textbf{FLOPs}  & \textbf{Params} & \textbf{Runtime}& \textbf{mIOU}\\ \hline
BiSeNet \cite{yu2018bisenet}  & 3.63G & 311K  & \textbf{3.24 ms} & 74.8\\
DANet \cite{fu2019dual}  & 1.01G & 82.24 K & 17.62 ms & 76.3 \\
GALDNet \cite{li2019global}  & 1.01G & 65.34K  & 14.28 ms & 76.1 \\
ANNNet \cite{zhu2019asymmetric}  & 0.82G & 42.24K  & 8.35 ms & 76.4 \\
GALD+SPP+CLO & \textbf{0.024} & \textbf{12.29K}  & 3.48 ms & \textbf{76.6} \\ \hline
\end{tabular}
\caption{Comparative study of different attention modules. 
%Incorporating SPP \cite{zhao2017pyramid} and cheap linear operations \cite{han2020ghostnet} in GALDNet \cite{li2019global} makes the attention module suitable for real-time applications. 
FLOPs, Params and Runtime correspond to the attention modules and not the overall architecture.}
\label{table:pam-cap}
\end{table}

\paragraph{\textbf{{Backbone Choice}}}

A lot of previous real-time semantic segmentation architectures \cite{yu2018bisenet, orsic2019defense, zhuang2019shelfnet, zhao2018icnet} employ powerful feature extractors like ResNet-18 \cite{he2015deep}. Even though this choice is justified for accurate semantic segmentation, the implications on execution speed and computational complexity are profound. Hence, for effective comparison, we replace our MobileNetV3 backbone with ResNet-18 and study the outcomes (Table \ref{table:without_backbone}).
From the table we confirm that our segmentation head is still lighter, faster and more accurate as compared to both SwiftNet \cite{orsic2019defense} and BiSeNet \cite{yu2018bisenet}, even if we use an expensive feature extractor like ResNet-18. Furthermore, the comparison between Ours-R18 and Ours-MV3 from Table \ref{table:without_backbone} reveals that the computational overheads added by ResNet-18 are larger as compared to MobileNetV3-Small even though they both provide similar mIOU scores. 
%Since the segmentation head is light, using a heavy backbone and limiting the capabilities of the overall model in terms of inference, does not seem to be optimal.

\begin{table}[ht]
\centering
\begin{tabular}{lcccccccc}
\hline
\textbf{Model}& \textbf{mIOU} & \textbf{Flops}  & \textbf{Params} & \textbf{FPS}\\ \hline
BiseNet-R18 \cite{yu2018bisenet}& 74.8 & 103.72G& 12.89M & 47.20\\
SwiftNet-R18 \cite{orsic2019defense} & 75.4 & 103.37G& 11.80M & 45.40 \\ \hline 
Ours-R18 & 76.7 & 66.41G & 9.19M& 54.50  \\
Ours-MV3 & 76.6 & 12.03G & 2.64M& 66.50  \\ \hline
\end{tabular}
\caption{Complexity comparison between our approach and the current state-of-the-art with different backbones. R18 and MV3 stand for ResNet-18 and MobileNetV3-Small ($1.\times$) respectively. %CABiNet-R18 offers a 35\% reduction in computations, with comparable mIOU along-with a 16\% reduction in the overall inference time. Both the approaches \cite{orsic2019defense, yu2018bisenet} use ResNet-18 as the primary feature extractor.
}
\label{table:without_backbone}
\end{table}

\paragraph{\textbf{{Feature Fusion Module}}}

Several fusion techniques have been suggested in the literature and designing the right one has significant impacts on the final outcome. Consider Table \ref{table:ffm_variants} for a quantitative comparison between the various fusion techniques. Feature concatenation with weighted attention and bottlenecks provides the best mIOU-FLOPs balance out of all the variants.
\begin{table}[ht]
\centering
\begin{tabular}{lcccccccc}
\hline
\textbf{Fusion Style}& \textbf{mIOU}  & \textbf{FLOPs} \\ \hline
Feature Addition \cite{poudel2018contextnet}& 73.2 & 0.5G \\
Feature Concatenation w/o AW  \cite{poudel2019fast} & 74.5 &  0.8G \\
Feature Concatenation w AW  \cite{yu2018bisenet}& 76.7 &  1.8G \\
Feature Concatenation w AW + Bottlenecks & 76.6 &  0.9G \\ \hline
\end{tabular}
\caption{
Comparative study of different fusion modules.
AW stands for attention weight based fusion. Addition of bottlenecks indicates the relative decrease in computations without significant impacts on mIOU.}
\label{table:ffm_variants}
\end{table}

\subsubsection{Results on Embedded Device}
\label{sec:jetson}

Inference on full scale GPUs (Titan X or RTX20 series) is unlikely to provide a real-world analysis, as autonomous vehicles, UAVs and UGVs are more likely to have low-power consumption modules with limited memory. 
In order to create effective comparisons, we further benchmark recent real-time semantic segmentation models on an embedded device Jetson Xavier NX, a small form factor system-on-module.  The device has a 384-core NVIDIA Volta GPU with 48 Tensor Cores and a 6-core NVIDIA Carmel ARM 64-bit CPU.  Results on the full resolution Cityscapes images are shown in the last column of Table~\ref{table:citys}.
Our model has the highest mIOU, and meantime remains relatively fast at 8.21 FPS.
Compared with Fast-SCNN \cite{poudel2019fast}, our method decreases at the speed of 3.28 FPS, but gains 7.7\% mIOU on the Cityscapes test set.  

We further experiment with our algorithm and others on multiple resolutions to demonstrate the efficiency.
% The mIOU scores have not been provided in Table \ref{table:jetson} as the models were trained and evaluated on different sizes and not all official implementations were available to evaluate. 
Table \ref{table:jetson} provides the system execution speed, memory footprints and GFLOP, counting on $\frac{1}{4}$ and $\frac{1}{16}$ of the full resolution Cityscapes images ($2048\times 1024$).
Note that, for full resolution image inference (1024$\times$2048) in Table~\ref{table:citys}, we employ the maximum power mode (15W, all 6 cores). However, for the smaller resolution,  we use a lower mode (10W, only 4 cores) because implementing semantic segmentation on lower resolutions is likely to imply that there could be more processes running, and hence considering the usage of other cores for other threads, we utilize only 4. The execution speed is affected by the number of processors involved in computations.

\begin{table}[http]
\centering
%\resizebox{}{}{}
\resizebox{1.01\textwidth}{!}{  
\begin{tabular}{l|cccl|l|cccl}
\hline
\textbf{Model}& \multicolumn{9}{c}{\textbf{Input Size}} \\ \hline
& \multicolumn{4}{c|}{\textbf{256x512}}&  & \multicolumn{4}{c}{\textbf{512x1024}}  \\ \hline
& \textbf{Memory} & \textbf{MAdd} & \textbf{FLOPs}& \multicolumn{1}{c|}{\textbf{FPS}}  &  & \textbf{Memory} & \textbf{MAdd} & \textbf{FLOPs}& \multicolumn{1}{c}{\textbf{FPS}}  \\ \hline
%ERFNet  & 201.50MB& 13.37G & 6.41G& 19.02 &  & 806.00MB& 53.47G & 25.65G & 5.20  \\ 
%ICNet& 68.40MB& 10.15G & 5.06G& 8.49  &  & 273.62MB& 40.61G & 20.26G & 3.38  \\ 
%DABNet  & 205.47MB& 5.18G& 2.62G& 41.06 &  & 821.88MB& 20.71G & 10.22G & 10.95 \\ 
CGNet \cite{wu2018cgnet} & 195.96MB& 3.44G& 1.69G& 25.77 &  & 786.75MB& 13.75G & 6.76G& 11.43 \\ 
ContextNet \cite{poudel2018contextnet}& 88.65MB& 869.38M& 419.61M& 60.97 &  & 356.44MB& 3.49G& 1.68G& 36.77 \\ 
SINet \cite{park2020sinet} & 42.00MB& 187.02M& 74.98M & 28.28 &  & 168.00MB& 784.04M& 299.90M& 24.78 \\ 
Fast-SCNN \cite{poudel2019fast}& 77.37MB& 865.74M& 419.78M& 69.64 &  & 309.71MB& 3.46G& 1.68G& 42.34 \\ 
DFANet \cite{li2019dfanet} & 111.16MB& 1.92G& 955.16M& 15.25 &  & 444.55MB& 7.67G& 3.82G& 13.82 \\ 
ShelfNet \cite{zhuang2019shelfnet}& 72.39MB& 11.74G & 5.86G& 26.97 &  & 289.53MB& 46.97G & 23.42G & 7.30  \\ 
SwiftNet \cite{orsic2019defense}& 104.66MB& 12.99G & 6.47G& 26.19 &  & 418.06MB& 51.95G & 25.89G & 9.14  \\ 
BiseNet \cite{yu2018bisenet}& 121.35MB& 13.01G & 6.48G& 28.07 &  & 418.36MB& 52.05G & 25.93G & 8.84  \\ 
%BiseNetV2 \cite{yu2020bisenet}& 174.06MB& 12.98G & 6.46G& 26.66 &  & 511.56MB& 50.25G & 24.93G & 8.12  \\ 
\hline
Ours & 61.23MB& 1.03G& 502.22M& 45.55 &  & 244.82MB& 4.10G& 2.01G& 35.72 \\ \hline
\end{tabular}
}
\caption{
Comparison with state-of-the-art on the Cityscapes dataset \cite{Cordts2016Cityscapes}, with respect to the system execution speed, memory footprints, and GFLOP, counting on $\frac{1}{4}$ and $\frac{1}{16}$ of the full resolution Cityscapes images. 
Note that FPS in this table indicates the average model run-times on a single Jetson Xavier NX. 
}
%\caption{Jetson Xavier NX has 6 modes of operation, depending on the power consumption and the number of cores utilized. For full resolution testing (1024$\times$2048), we employ the maximum power mode (15W, all 6 cores). However, for the smaller resolutions (512$\times$1024 and 256$\times$512) we use a lower mode (10W, only 4 cores) to establish an effective comparison between the possible use-cases. For instance, implementing semantic segmentation on lower resolutions is likely to imply that there could be more processes running, and hence considering the usage of other cores for other threads, we utilize only 4. The execution speed is affected by the number of processors involved in computations.}
% \caption{For full resolution testing (1024$\times$2048) in Table~\ref{table:citys}, we employ the maximum power mode (15W, all 6 cores). However, for the smaller resolution,  we use a lower mode (10W, only 4 cores) because implementing semantic segmentation on lower resolutions is likely to imply that there could be more processes running, and hence considering the usage of other cores for other threads, we utilize only 4. The execution speed is affected by the number of processors involved in computations.}
\label{table:jetson}
\end{table}

% \begin{table}[h!]
% \centering
% \small
% \begin{tabular}{lcccllcccl}
% \hline
% \textbf{Model}& \textbf{Memory} & \textbf{MAdd} & \textbf{FLOPs}& \textbf{FPS}  \\ \hline
% BiseNetV2 \cite{yu2020bisenet}   & 511.56MB& 50.25G & 24.93G & 8.12  \\ 
% SwiftNet \cite{orsic2019defense} & 418.06MB& 51.95G & 25.89G & 9.14  \\ 
% ShelfNet \cite{zhuang2019shelfnet} & 289.53MB& 46.97G & 23.42G & 7.30  \\ 
% ContextNet \cite{poudel2018contextnet} & 356.44MB& 3.49G& 1.68G& 36.77 \\ 
% Fast-SCNN  \cite{poudel2019fast} & 309.71MB& 3.46G& 1.68G& 42.34 \\ \hline
% CABiNet (Ours)  & 244.82MB& 4.10G& 2.01G& 35.72 \\ \hline
% \end{tabular}
% \caption{For full resolution testing (1024$\times$2048) in Table \ref{table:citys}, we employ the maximum power mode (15W, all 6 cores). However, for the smaller resolution (512$\times$1024) we use a lower mode (10W, only 4 cores) because implementing semantic segmentation on lower resolutions is likely to imply that there could be more processes running, and hence considering the usage of other cores for other threads, we utilize only 4. The execution speed is affected by the number of processors involved in computations.}
% \label{table:jetson}
% \end{table}

\subsection{Experiments on UAVid dataset}

\subsubsection{Dataset}
We also evaluate our Context Aggregation Network by another semantic segmentation dataset UAVid \cite{lyu2020uavid}.
UAVid dataset is a high-resolution UAV semantic segmentation dataset focusing on street scenes.
The dataset contains a total of 420  images out of which 200 are for training, 70 for validation and the remaining 150 for testing.  The resolution of the images is 3840$\times$2160.
UAVid \cite{lyu2020uavid} is a challenging benchmark due to the large resolution of images, large scale variation, and  complexities in the scenes. 
We use the same hyperparameters and data augmentation as those for experiments on Cityscapes dataset, except the initial learning rate as $5e^{-5}$, batch size as 3, and training iterations as 240k.
We split the image into 4 equal quarters of 1920$\times$1080 during training.

% Hera again, we do not use the validation data during training. The validation data is only used for hyperparamter tuning post training. Testing is done on the UAVid official servers by submitting the predictions from the different models.

\subsubsection{Results}

Table \ref{table:uavid} provides a detailed quantitative results of our model and the previous state-of-the-art method - MSD \cite{lyu2020uavid}.
Our model outperforms MSD by a large margin of 6.5\%, while maintaining an execution speed of 15 FPS with the full resolution image on a single GPU  RTX 2080Ti. 
Furthermore, we also train several recent real-time semantic segmentation models on UAVid dataset and report the results on UAVid test set from the official server \footnote{https://uavid.nl/} in Table \ref{table:uavid}.
However, we do not benchmark with this dataset on the Jetson Xavier NX, since it is not powerful enough. 
% Please note that for training ShelfNet \cite{zhuang2019shelfnet}, we adopt the same strategy mentioned in \cite{lyu2020uavid}, as the architecture functions with only fixed input-sizes which are multiples of 256. All models were trained on a batch-size of 3, for 50\% larger iterations than were originally proposed in each. $-$ indicates that the FPS of the algorithm could not be confirmed.
Our model achieves the best IOU score on six out of eight classes, and the best mean IOU with 2\% gain over BiSeNet \cite{yu2018bisenet} while being 4 FPS faster. Although Fast-SCNN \cite{poudel2019fast} is 18.7 FPS faster than ours, their mIOU is worse by a large margin of 17.6\%.      

\begin{table}[http]
\centering
\resizebox{1.01\textwidth}{!}{  
\begin{tabular}{lcccccccccl}
\hline
\textbf{Model} & \multicolumn{1}{l}{\textbf{Building}} & \multicolumn{1}{l}{\textbf{Tree}} & \multicolumn{1}{l}{\textbf{Clutter}} & \multicolumn{1}{l}{\textbf{Road}} & \multicolumn{1}{l}{\textbf{Vegetation}} & \multicolumn{1}{l}{\textbf{Static Car}} & \multicolumn{1}{l}{\textbf{Moving Car}} & \multicolumn{1}{l}{\textbf{Human}} & \multicolumn{1}{l}{\textbf{mIOU}} & \textbf{FPS} \\ \hline
MSD \cite{lyu2020uavid} & 79.8  & 74.5  & 57.0 & \textbf{74.0} & 55.9 & 32.1 & \textbf{62.9} & 19.7 & 57.0  & 1.00 \\ \hline
Fast-SCNN \cite{poudel2019fast} & 75.7  & 71.5  & 44.2 & 61.6 & 43.4& 19.5 & 51.6 & 0.0 &45.9&\textbf{33.84} \\
ShelfNet \cite{zhuang2019shelfnet} & 76.9  & 73.2  & 44.1 & 61.4 & 43.4& 21.0 & 52.6  & 3.6  &47.0 &9.65\\
SwiftNet \cite{orsic2019defense} & 85.3  & 78.2  & 64.1 & 61.5  & 76.4& 62.1 & 51.1 & 15.7& 61.1  &11.84\\
BiSeNet \cite{yu2018bisenet}  & 85.7  & 78.3  & 64.7 & 61.1  & 77.3& 63.4 & 48.6 & 17.5& 61.5  &  11.08 \\ \hline
Ours & \textbf{86.6}  & \textbf{79.3}  & \textbf{66.0} & 62.1  & \textbf{78.1}& \textbf{68.3} & 47.8 & \textbf{19.9}& \textbf{63.5}  &  15.14\\ \hline
\end{tabular}
}
\caption{
Quantitative comparisons between other methods and our model on the UAVid test set \cite{lyu2020uavid}.
The first eight columns show IOU scores for eight classes. 
The last two columns list mean IOU score and the execution speed. The boldface text means the highest value in the column.}
\label{table:uavid}
\end{table}

\begin{figure}[http]
\centering
	\includegraphics[width=\textwidth]{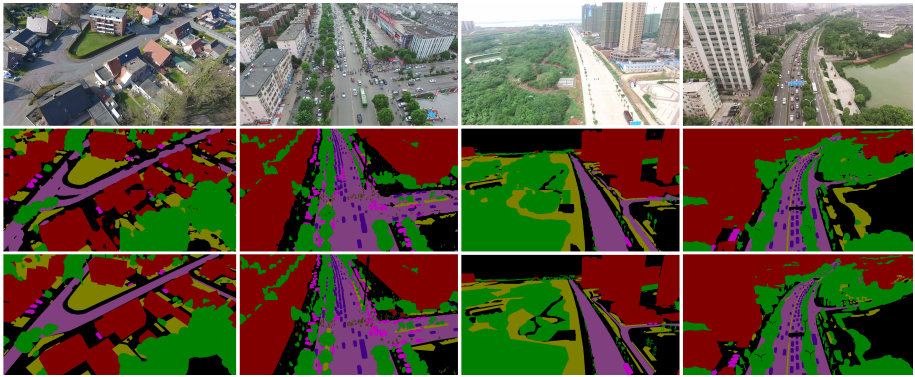}
			\caption{Semantic segmentation results on the UAVid \cite{lyu2020uavid} validation set. First row consists of the input RGB images. Second row contains the predictions from our model and the third row shows the ground truth of the input images.}
	\label{fig:cabinet-uavid-val}
\end{figure}

\begin{figure}[http]
\centering
	\includegraphics[width=\textwidth]{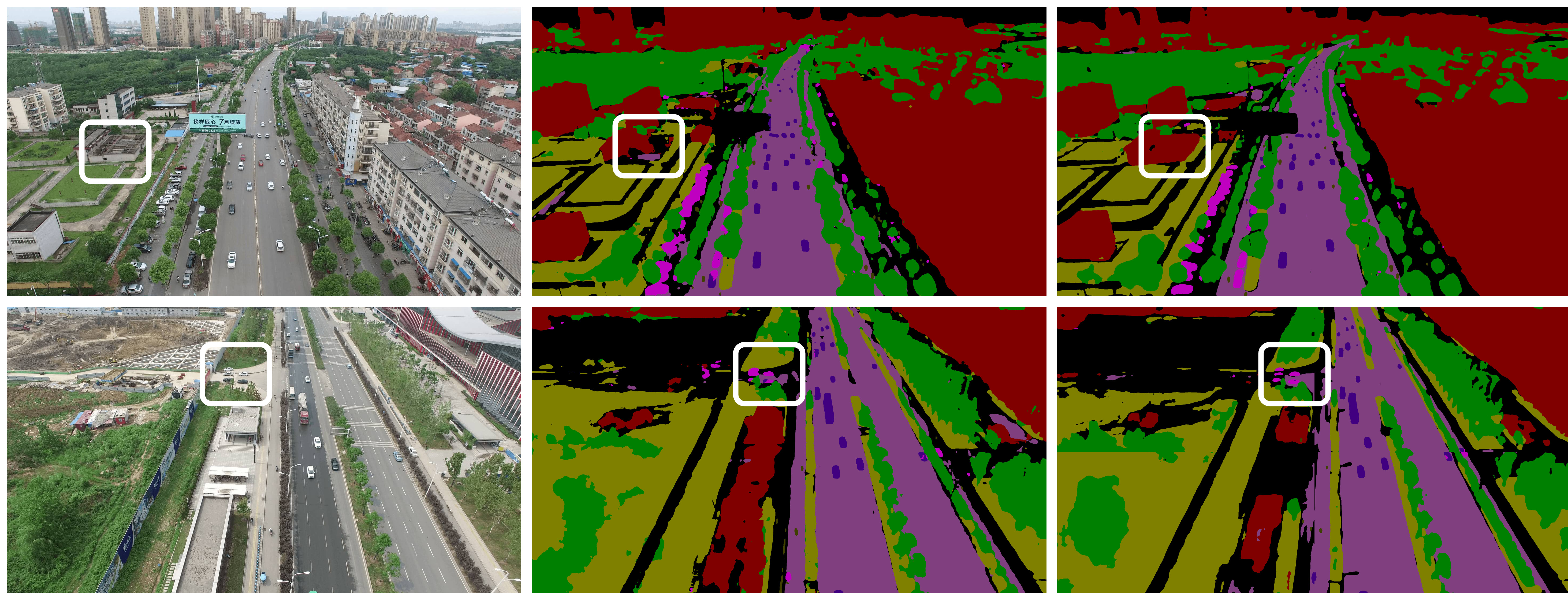}
			\caption{Comparative segmentation results from the UAVid \cite{lyu2020uavid} test dataset. First column shows the input RGB images, second column depicts the outputs of MSD \cite{lyu2020uavid} and the third column shows the predictions of our model. White boxes highlight the regions of improvement.}
	\label{fig:cabinet-uavid-test}
\end{figure}

Qualitative results on UAVid validation set and test set are shown in Fig. \ref{fig:cabinet-uavid-val} and Fig. \ref{fig:cabinet-uavid-test}, respectively.
As it can be seen from Fig. \ref{fig:cabinet-uavid-test}, compared with MSD \cite{lyu2020uavid}, our model does not suffer from local or global inconsistencies, thereby effectively capturing the cues to scene semantics. 

\section{Conclusions}
\label{sec:con}
In this paper, we have developed a light-weight Context Aggregation Network to address the challenge of real-time semantic segmentation with improved inference speeds and reduced computational expenses. 
Building upon the  dual-branch architectures for high-speed semantic segmentation, we design a cheap high resolution branch for effective spatial detailing and a context branch with light-weight versions of global aggregation and local distribution blocks, potent to capture both long-range and local contextual dependencies required for accurate semantic segmentation, with low computational overheads.
Our proposed approach is trained end-to-end.
We verify the advantages of our proposed method by comprehensive ablation experiments on
Cityscapes dataset. 
When comparing to other state-of-the-art models, our model achieves 76.6\% and 75.9\% mIOU on Cityscapes validation and test sets respectively, at 76 FPS on an NVIDIA RTX 2080Ti and 8 FPS on a Jetson Xavier NX.
Furthermore, we conduct experiments on the UAVid dataset and our model produces more
accurate pixelwise semantic predictions in terms of mIOU (63.5\%) with high execution speed (15 FPS),
compared to the baseline and other leading models. 
This further demonstrates the advantage of our proposed method.
For future work, we will extend the current approach to address low-latency instance segmentation.

\bibliography{bibliography}

\end{document}